\documentclass[journal]{IEEEtran}

\usepackage{cite}

\usepackage{algorithmic}
\usepackage{algorithm2e}

\usepackage{color}
\usepackage{placeins}
\usepackage{float}
\usepackage{tabularx,colortbl}

\usepackage{subfigure}

\usepackage[cmex10]{amsmath}
\usepackage[pdftex]{graphicx}
\usepackage{amsfonts}

\usepackage{xcolor}

\begin{document}
%
\title{Detecting Breast Cancer using a Compressive Sensing Unmixing Algorithm}

\author{\IEEEauthorblockN{Richard Obermeier and Jose~Angel~Martinez-Lorenzo}}



\maketitle

\begin{abstract}
Traditional breast cancer imaging methods using microwave Nearfield Radar Imaging (NRI) seek to recover the complex permittivity of the tissues at each voxel in the imaging region. This approach is suboptimal, in that it does not directly consider the permittivity values that healthy and cancerous breast tissues typically have. In this paper, we describe a novel unmixing algorithm for detecting breast cancer. In this approach, the breast tissue is separated into three components, low water content (LWC), high water content (HWC), and cancerous tissues, and the goal of the optimization procedure is to recover the mixture proportions for each component. By utilizing this approach in a hybrid DBT / NRI system, the unmixing reconstruction process can be posed as a sparse recovery problem, such that compressive sensing (CS) techniques can be employed. A numerical analysis is performed, which demonstrates that cancerous lesions can be detected from their mixture proportion under the appropriate conditions. 
\end{abstract}


%
\IEEEpeerreviewmaketitle

\vspace{7pt}
\section{Introduction}
\label{sec:intro}
A recent report by the Center for Disease Control and Prevention \cite{Health2013} states that breast cancer is the most common type of cancer among women, with a rate of $118.7$ cases per $100,000$ women, and that it is the second deadliest type of cancer among women, with a mortality rate of $21.9$ deaths per $100,000$ women. It is well known that the detection of breast cancer in its early stages can greatly improve a woman's chance for survival, as the lesions tend to be smaller and are less likely to have spread from the breast than more developed cancer. Although small cancers near the surface of the breast can be detected by means of a clinical breast exam (CBE), cancers deep within the breast can only be detected through non-invasive imaging.

X-ray based technologies such as Conventional Mammography (CM) and Digital Breast Tomosynthesis (DBT) are most often used to detect cancerous lesions within the breast. Unfortunately, these systems both suffer from the small radiological contrast between health breast tissue and cancerous tissue, which is on the order of $1\%$. As a result, these technologies tend to produce a large number of false positives when used for early detection. Nearfield Radar Imaging (NRI) is a less common technology for breast cancer detection that uses non-ionizing microwave radiation to excite the breast. NRI is an appealing technology for breast cancer detection because the contrast between healthy breast tissue and cancerous tissue is on the order of $10\%$ at microwave frequencies \cite{Lazebnik2007}. Unfortunately, standalone NRI systems typically struggle to accurately detect cancerous lesions due to the heterogeneous distribution of tissues within the breast.

Recent papers \cite{MartinezLorenzo2013,obermeier2015a,obermeier2016compressive} have introduced the concept of using a hybrid DBT / NRI system for breast cancer detection. In the hybrid system, the DBT reconstruction is used in order to form a prior distribution of tissues in the breast that can be used as a starting point for the NRI inversion process. By applying the Born Approximation, one can formulate a linear sensing problem for the complex permittivity of the breast tissues \cite{obermeier2015a,obermeier2016compressive}. In addition, the reconstruction process can be formulated as a sparse recovery problem, such that novel compressive sensing (CS) techniques \cite{Candes2006,Candes2006a,Donoho2006} can be applied. Although this technique has been applied with some success, it has one critical shortcoming in that it does not accurately model the dispersive nature of the the breast tissues.

Unmixing algorithms \cite{keshava2002spectral,keshava2003survey} have been widely applied to hyperspectral sensing applications. In these applications, the composition of each pixel is expressed as a mixture of several different materials. Consequently, the goal of the unmixing algorithms is to reconstruct the mixture proportions of each material at each pixel. Unmixing algorithms have been used to solve many types of applications, including nonlinear \cite{dobigeon2014nonlinear} and sparse recovery problems \cite{iordache2011sparse,bioucas2010alternating,bioucas2009variable}. In this work, we propose the use of a compressed sensing unmixing algorithm for breast cancer detection in the hybrid DBT / NRI system. Like the hyperspectral imaging applications, the composition of each voxel in breast imaging applications can be decomposed into a mixture of three different materials: high water content (HWC) healthy breast tissue, low water content (LWC) healthy breast tissue, and cancerous lesions. The goal of the unmixing reconstruction algorithm, then, is to recover the mixture proportions of the three material types at each voxel.

The remainder of this paper is organized as follows. In Section \ref{sec:nonlinear}, we describe a general formulation for unmixing problems as they apply to traditional nonlinear inverse problems (i.e. electromagnetics, acoustics, etc.). In Section \ref{sec:linear}, we describe a simpler linearized unmixing formulation based upon the Born Approximation, which could in turn be used to iteratively solve the nonlinear problem. In Section \ref{sec:cs}, we describe a compressed sensing unmixing algorithm that can be used by the hybrid DBT / NRI system. We also describe an accelerated gradient Augmented Lagrangian method for solving this unmixing problem. In Section \ref{sec:results}, we present numerical results that demonstrate the imaging capabilities of the unmixing algorithm. Finally, we conclude the paper in Section \ref{sec:conclusion} by discussing some possible extensions to the method.

\vspace{7pt}
\section{General Nonlinear Unmixing Problem}
\label{sec:nonlinear}
Suppose than an object of interest is excited by a sensing system whose measurements are governed by the relationship $y = f(x)$, where $x$ is related to the constitutive parameters of the sensing modality (i.e. $\epsilon$, $\sigma$, and $\mu$ for electromagnetics, $\rho$, and $\kappa$ for acoustics, etc.), and $f(\cdot)$ is a possibly nonlinear function of the constitutive parameters that describes the measurement process. The goal is to recover the vector $x$ from the set of measurements $y$. Without any prior knowledge about the object of interest, the unknown vector $x$ can take any value permitted by the laws of physics. In this case, it is often difficult to accurately reconstruct $x$. However, suppose that the object is constructed from a mixture of $R$ different materials, and let $z_r$ denote the fraction of the $r$-th material at each position in the imaging region. Suppose also that there exists a possibly nonlinear model for computing $x$ from $z_r$, i.e. $x = h(z_1,z_2,\ldots,z_R$). The goal in this problem is to recover the mixture values $z_1,z_2,\ldots,z_R$ from the the measurement vector $y$. 

In order to accurately recover the mixture proportions, one must use a suitable inversion algorithm. In practice, the number of measurements $M$ is less than the number of unknown variables $N R$, and so it is not possible to recover a unique set of mixture proportions. Furthermore, the measurements are typically corrupted by noise, and so the recovery process should not seek out an exact solution to the nonlinear system. Finally, the inversion algorithm must ensure that the final result represents a valid mixture proportion. Taking all of these qualities into consideration, we propose the following optimization program as a solution to the unmixing problem:
\begin{alignat}{3}
& \underset{z_1, z_2, \ldots, z_R}{\text{minimize }}~~ & & g(z_1, z_2, \ldots, z_R) \label{eq:nl1} \\
&\text{subject to } &  & \|y - f(h(z_1, z_2, \ldots, z_R))\|_{\ell_2} \le \delta \nonumber \\
& & & z_r \succeq 0_N, ~r = 1, 2, \ldots, R \nonumber \\
& & & \sum_{r=1}^R z_r = 1_N \nonumber \\
& & & \alpha_{mn}\left(z_m \odot z_n\right) = 0_N ~~\forall m \neq n \nonumber
\end{alignat}
where $0_N$ and $1_N$ are column vectors containing $N$ zeros and ones respectively, and $\odot$ represents Hadamar (element-wise) product. In Eq. \ref{eq:nl1}, the objective function $g(\cdot)$ acts as a regularizer for the mixture proportions and is derived from prior knowledge, and $\delta$ represents an estimate of the error in the measurement vector. The positivity and linear equality constraints ensure that the solution vectors represent valid mixture proportions (non-negative and sum to one), and the nonlinear equality constraint ensures that certain mixtures cannot coexist, when applicable. For example, if a given mixture can only have a component of $z_m$ or a component of $z_n$, but not both simultaneously, then setting $\alpha_{mn} \neq 0$ in the optimization program will enforce this constraint. If a mixture of $z_m$ and $z_n$ is allowed, then setting $\alpha_{mn} = 0$ will permit that possibility.  

\vspace{7pt}
\section{Simplified Linear Unmixing Problem}
\label{sec:linear}
The nonlinear problem of Eq. \ref{eq:nl1} poses a number of challenges for practical sensing applications. First, the constraint $\alpha_{mn}\left(z_m \odot z_n\right) = 0_N$ for $\alpha_{mn} \neq 0$ ensures that the optimization program is nonconvex and combinatorial in nature. Second, if either $f(\cdot)$ or $h(\cdot)$ are nonlinear, then the quadratic error constraint becomes nonconvex, increasing the difficulty of the problem even further. Finally, the optimization procedure can be very computationally expensive if a complicated forward model solver is required in order to evaluate $f(\cdot)$; this is the case for electromagnetic sensing systems. 

In order to simplify the optimization procedure, we propose the following two approximations. First, the constraints $\alpha_{mn}\left(z_m \odot z_n\right) = 0_N$ are discarded. As a result of this approximation, it is possible that the solutions vectors may violate these constraints. Nevertheless, this approximation is necessary in order reduce the computational complexity of the problem. Second, the quadratic constraint is simplified using a linear approximation to $f(h(\cdot))$ about a prior set of solution vectors $v_1, v_2, \ldots, v_R$. Formally, we express this linearization as follows:
\begin{align}
f(h(z_1, z_2, \ldots, z_R)) &\approx f(h(v_1, v_2, \ldots, v_R)) \nonumber \\
&+ \sum_{r=1}^R F H_{r} (z_r -v_r)
\end{align}
where $F = \frac{\partial}{\partial x} f(x) |_{x = h(v_1, v_2, \ldots, v_R)}$ and $H_r = \frac{\partial}{\partial z_r} h(z_1, z_2, \ldots, z_R) |_{z_r = v_r}$. To simplify notation, we introduce two substitutions,  $A_r = F H_r$ and $\hat{y} = y - f(h(v_1, v_2, \ldots, v_R)) + \sum_{r=1}^R A_r v_r$. With these simplifications, the simplified unmixing problem can be expressed as follows:
\begin{alignat}{3}
& \underset{z_1, z_2, \ldots, z_R}{\text{minimize }}~~ & & g(z_1, z_2, \ldots, z_R) \label{eq:lin1} \\
&\text{subject to } &  & \|\hat{y} - \sum_{r=1}^R A_r z_r\|_{\ell_2} \le \delta \nonumber \\
& & & z_r \succeq 0_N, ~r = 1, 2, \ldots, R \nonumber \\
& & & \sum_{r=1}^R z_r = 1_N \nonumber
\end{alignat}

\vspace{7pt}
\section{Compressed Sensing Unmixing for the Hybrid DBT / NRI System}
\label{sec:cs}
\subsection{Mixture Parameters}
When applied to breast cancer detection, the unmixing problem can be expressed in terms of three mixture parameters $z_1$, $z_2$, and $z_3$, for the HWC breast tissue, LWC breast tissue, and cancerous tissue. In this formulation, the material elements can more coarsely separated into two groups, healthy tissue and cancerous tissue. Consequently, one should enforce $\alpha_{13} \neq 0$ and $\alpha_{23} \neq 0$ in Eq. \ref{eq:nl1}. However, following the simplifications of Section \ref{sec:linear}, this constraint will be ignored in order to decrease the computational complexity of the optimization procedure. 

To complete the unmixing formulation, we must define the functions $f(\cdot)$ and $h(\cdot)$. The measurement function $f(\cdot)$ is, of course, derived from Maxwell's equations. Assuming that the NRI system directly measures the electric field vector, the measurement function $f(\cdot)$ can be derived from the following relationship:
\begin{align}
\mathbf{E}(\mathbf{r}_r,\omega) &=  \jmath\omega\int \mathbf{G}(\mathbf{r}_r, \mathbf{r}', \omega; \epsilon(\mathbf{r}',\omega)) \mathbf{I}(\mathbf{r}',\omega) d\mathbf{r}'  \label{eq:em}
\end{align}
where $\mathbf{r}_r$ is the receiver position, $\omega$ is the radial frequency, $\mathbf{E}(\mathbf{r}_r,\omega)$ is the total electric field vector, $\epsilon(\mathbf{r},\omega)$ is the complex permittivity, $\mathbf{G}(\mathbf{r}_r, \mathbf{r}', \omega; \epsilon(\mathbf{r}',\omega))$ is the dyadic Green's function of the breast, and $\mathbf{I}(\mathbf{r}',\omega)$ is the current source distribution for the transmitting antennas. The Green's function $\mathbf{G}(\mathbf{r}_r, \mathbf{r}', \omega; \epsilon(\mathbf{r}',\omega))$ is a nonlinear function of the permittivity vector, and must be computed using a forward model solver such as the finite differences in the frequency domain (FDFD) method \cite{Rappaport2001}. As a result, it would be prohibitively expensive to implement the full nonlinear method of Eq. \ref{eq:nl1} in practice. The function $h(\cdot)$ is derived from the clinical analysis performed by Lazebnik et. al \cite{Lazebnik2007}. In their work, Lazebnik et. al generated Cole-Cole models for the complex permittivity of HWC, LWC, and cancerous tissues. A linear model is used to compute the permittivity of an arbitrary mixture of the three tissue types in the unmixing procedure. This can be explicitly written as $\epsilon = \sum_{r=1}^3 z_r\epsilon_r$. 

To compute the initial starting points $v_1$, $v_2$, and $v_3$, the DBT reconstruction is segmented under the assumption that the breast contains only healthy tissue. In this approach, the water content content is extracted from the DBT image such that $v_2 = 1_N - v_1$ and $v_3 = 0_N$. This approach is possible because the X-ray attenuation coefficient is intimately related to the water content of the tissue. The starting points $v_1$, $v_2$, and $v_3$ are in turn used to compute the Jacobian matrices $A_r$ and the adjusted measurements $\hat{y}$.

\subsection{Compressive Sensing}
If the DBT image is accurately segmented, then the mixture proportions will be correct everywhere except at the positions of the cancerous lesions. Furthermore, since cancerous lesions tend to be localized to small regions of the breast, the imaging problem can be posed as a sparse recovery problem. This motivates the use of novel compressive sensing (CS) techniques \cite{Candes2006,Candes2006a,Donoho2006}. CS theory is a novel signal processing paradigm, which states that sparse signals can be recovered using a small number of linear measurements. For a traditional CS problem, sparse solution vectors $x$ can be recovered from a set of linear measurements $y$ by solving the following $\ell_1$-norm minimization problem:
\begin{alignat}{3}
& \underset{{x}}{\text{minimize }}~~ & &\|{x}\|_{\ell_1} \label{eq:ell1norm-1} \\
&\text{subject to } &  &\|{A}{x} -{y}\|_{\ell_2} \le \eta \nonumber
\end{alignat}
provided that $A$ obeys a Restricted Isometry Property, which can be defined as follows. For a fixed sparsity level $S$, the restricted isometry constant $\delta_S$ is the smallest positive constant such that
 \begin{align}
 (1-\delta_S)\|x\|_{\ell_2}^2 \le \|Ax\|_{\ell_2}^2 \le (1+\delta_S)\|x\|_{\ell_2}^2 \label{eq:ric}
 \end{align}
is satisfied for all vectors with $\|x\|_{\ell_0} \le S$, where the ``$\ell_0-$norm'' measures the number of non-zero elements in the vector. In other words, the restricted isometry constant $\delta_S$ establishes bounds for the singular values of submatrices obtained by selecting any $S$ columns from the complete sensing matrix $A$. Candes showed in \cite{candes2008} that the distance between the optimal solution $x^*$ to Eq. \ref{eq:ell1norm-1} and the true sparse vector $x_t$ is bounded according to
\begin{align}
\|x^*-x_t\|_{\ell_2} \le C \eta \label{eq:bound}
\end{align}
provided that the restricted isometry constant $\delta_{2S} < \sqrt{2} - 1$.

The additional linear constraints of Eq. \ref{eq:lin1} prevent the unmixing problem from being posed in the traditional CS format of Eq. \ref{eq:ell1norm-1}. Nevertheless, CS techniques can be applied to the unmixing problem with some modifications. As previously stated, the mixture proportions should differ from the initial values segmented from the DBT image at a small number of locations. As a result, we argue that the unmixing problem should seek solution vectors where $z_r - v_r$ is sparse. Therefore, we propose the following CS optimization program for the unmixing problem:
\begin{alignat}{3}
& \underset{z_1, z_2, z_3}{\text{minimize }}~~ & & \sum_{r=1}^3 \|z_r - v_r\|_{\ell_1} \label{eq:csunmix} \\
&\text{subject to } &  & \|\hat{y} - \sum_{r=1}^3 A_r z_r\|_{\ell_2} \le \delta \nonumber \\
& & & z_r \succeq 0_N, ~r = 1, 2, 3 \nonumber \\
& & & \sum_{r=1}^3 z_r = 1_N \nonumber
\end{alignat} 

\subsection{Solution using the Augmented Lagrangian}
As previously stated, the additional equality constraints in Eq. \ref{eq:csunmix} prevent traditional CS techniques from being used in order to solve the problem. In this section, we describe a method based upon the Augmented Lagrangian in order to solve Eq. \ref{eq:csunmix}. To simplify the notation, let us recast Eq. \ref{eq:csunmix} in terms of a single variable $z = \left(z_1^T, z_2^T, z_3^T\right)^T$ to represent the mixing parameters; doing so leads to Eq. \ref{eq:unmixB}:
\begin{alignat}{3}
& \underset{z}{\text{minimize }}~~ & & \|z - v\|_{\ell_1} \label{eq:unmixB}\\
&\text{subject to } &  & \|\hat{y} - Az\|_{\ell_2} \le \delta \nonumber \\
& & & z \succeq 0_{N_t} \nonumber \\
& & & D z = 1_{N} \nonumber
\end{alignat}
where $v = \left(v_1^T, v_2^T, v_3^T\right)^T$, $N_t = 3N$, $D = 1_3^T \otimes I_N$, $A = \left(A_1, A_2, A_3\right)$, and $\otimes$ is the Kronecker product. Using the auxiliary variables $w_1, w_2, w_3, w_4$, the constraints can be separated, leading to the equivalent optimization problem: 
\begin{alignat}{3}
& \underset{z,w_1, w_2, w_3, w_4}{\text{minimize }}~~ & & \|w_1\|_{\ell_1} + I_{Q_{\ell_2}}(w_2) + I_{\text{box}}(w_3) + I_{\text{D}}(w_4)  \nonumber\\
&\text{subject to } &  & w_1 = z - v \label{eq:unmixC} \\
& & & w_2 = Az - \hat{y} \nonumber \\                                   
& & & w_3 = z \nonumber \\                                        
& & & w_4 = z \nonumber
\end{alignat}
where $I_{Q_{\ell_2}}(\cdot)$ is the indicator function for the squared error set, $I_{\text{box}}(\cdot)$ is the indicator function for the positivity box constraint, and $I_{\text{D}}(\cdot)$ is the indicator function for the equality constraint. The method of multipliers, also known as the Augmented Lagrangian method \cite{nocedal2006numerical,Boyd2011a}, solves this problem by forming the unconstrained Augmented Lagrangian, which can be expressed as follows: 
\begin{align}
&\mathcal{L}_A(z,w_1, w_2, w_3, w_4,u_1, u_2, u_3, u_4) = \label{eq:auglag} \\
& \|w_1\|_{\ell_1} + I_{Q_{\ell_2}}(w_2) + I_{\text{box}}(w_3) + I_{\text{D}}(w_4) + \nonumber \\
&(\rho/2)\|w_1 - z + v + u_1\|_{\ell_2}^2 + (\rho/2)\|w_2 + \hat{y} - Az + u_2\|_{\ell_2}^2 + \nonumber \\
& (\rho/2)\|w_3 - z + u_3\|_{\ell_2}^2 + (\rho/2)\|w_4 - z + u_4\|_{\ell_2}^2 \nonumber
\end{align}
where $\rho$ is a positive scalar constant and $u_1$, $u_2$, $u_3$, and $u_4$ are the scaled dual variables. In its current form, Eq. \ref{eq:auglag} can be solved using the Alternating Direction Method of Multipliers (ADMM) \cite{Boyd2011a}. This method solves Eq. \ref{eq:auglag} by performing alternating minimizations of $z$, $w_1$, $w_2$, $w_3$, and $w_4$, before updating the dual variables using the method typically used in the method of multipliers. 

One drawback of using the ADMM for Eq. \ref{eq:auglag}, however, is the $z$ update step, which can be computationally expensive due to the large dimensionality of $A$. Instead, let us consider the more traditional Augmented Lagrangian method, in which the variables $z$, $w_1$, $w_2$, $w_3$, and $w_4$ are jointly optimized before the dual variables are updated. A close inspection of Eq. \ref{eq:auglag} reveals that the optimization terms can be separated into two groups: the smooth and Lipschitz continuously differentiable terms (the quadratic terms) and the non-smooth terms, which have closed-form proximal operators (this is demonstrated later). These qualities allow the Augmented Lagrangian subproblem of  Eq. \ref{eq:auglag} to be solved using accelerated proximal gradient techniques, which have a convergence rate of $\mathcal{O}(1/k^2)$. In this work, we propose the accelerated gradient method used by the fast iterative shrinkage thresholding algorithm (FISTA) \cite{Beck2009,Vandenberghe2014}. Algorithm \ref{alg:fista_auglag_sub} summarizes the use of FISTA to solve Eq. \ref{eq:auglag}. Note that the step size $t^{(k)}$ can be computed using an efficient line search method; see \cite{Beck2009,Vandenberghe2014} for details. Once a single iteration of Eq. \ref{eq:auglag} is solved, the dual variables are updated according to Eq. \ref{eq:u1} - \ref{eq:u4}.
\begin{algorithm}[t!]
Given $x^{(0)}, w_1^{(0)}, \ldots, w_4^{(0)}, u_1, \ldots, u_4$ \BlankLine
$x^{(0)} = x^{(-1)} = x^{(-2)}$ \BlankLine
$z_r^{(0)} = z_r^{(-1)} = z_r^{(-2)}~,~r=1,\ldots,4$ \BlankLine
\For{k = $1,2,3,\ldots$}{
Compute $\hat{z}^{(k)},  \hat{w}_1^{(k)}, \ldots, \hat{w}_4^{(k)}$ \BlankLine
\Indp $\hat{z}^{(k)} = z^{(k-1)} + \frac{k-2}{k+1}\left(z^{(k-1)} - z^{(k-2)}\right)$  \BlankLine
$\hat{w}_r^{(k)} = w_r^{(k-1)} + \frac{k-2}{k+1}\left(w_r^{(k-1)} - w_r^{(k-2)}\right)$ \BlankLine
\Indm Compute ${z}^{(k+1)},  {w}_1^{(k+1)}, \ldots, {w}_4^{(k+1)}$ \BlankLine
\Indp $z^{(k+1)} = \hat{z}^{(k)} +  t^{(k)} \rho \left(\hat{w}_1^{(k)} - \hat{z}^{(k)} + v + u_1\right) + t^{(k)} \rho A^H\left(\hat{w}_2^{(k)} + \hat{y} - A\hat{z}^{(k)} + u_2\right) + t^{(k)} \rho \left(\hat{w}_3^{(k)} - \hat{z}^{(k)} + u_3\right) + t^{(k)} \rho \left(\hat{w}_4^{(k)} - \hat{z}^{(k)} + u_4\right)$  \BlankLine
$w_1^{(k+1)} = \operatorname{prox}_{t^{(k)}\|\cdot\|_{\ell_1}}\left(\hat{w}_1^{(k)} - t^{(k)}\rho \left(\hat{w}_1^{(k)} - \hat{z}^{(k)} + v + u_1\right)\right)$ \BlankLine
$w_2^{(k+1)} = \operatorname{prox}_{t^{(k)}Q_{\ell_2}}\left(\hat{w}_2^{(k)} - t^{(k)}\rho \left(\hat{w}_2^{(k)} + \hat{y} - A\hat{z}^{(k)} + u_2\right)\right)$ \BlankLine
$w_3^{(k+1)} = \operatorname{prox}_{t^{(k)}\text{box}}\left(\hat{w}_3^{(k)} - t^{(k)}\rho \left(\hat{w}_3^{(k)} - \hat{z}^{(k)} + u_3\right)\right)$ \BlankLine
$w_4^{(k+1)} = \operatorname{prox}_{t^{(k)}D}\left(\hat{w}_4^{(k)} - t^{(k)}\rho \left(\hat{w}_4^{(k)} - \hat{z}^{(k)} + u_4\right)\right)$ \BlankLine
}
\caption{FISTA applied to the subproblem of Eq. \ref{eq:auglag}.}
\label{alg:fista_auglag_sub}
\end{algorithm}
\begin{align}
u_1^{(k+1)} &= u_1^{(k)} + \hat{w}_1^{(k)} - \hat{z}^{(k)} + v \label{eq:u1} \\
u_2^{(k+1)} &= u_2^{(k)} + \hat{w}_2^{(k)} + \hat{y} - A\hat{z}^{(k)}  \label{eq:u2} \\
u_3^{(k+1)} &= u_3^{(k)} + \hat{w}_3^{(k)} - \hat{z}^{(k)}  \label{eq:u3} \\
u_4^{(k+1)} &= u_4^{(k)} + \hat{w}_4^{(k)} - \hat{z}^{(k)}  \label{eq:u4}
\end{align}

In order to apply FISTA to Eq. \ref{eq:auglag}, we require expressions for the proximal operators of the $\ell_1$-norm and the indicator functions $I_{Q_{\ell_2}}(\cdot)$, $I_{\text{box}}(\cdot)$, and $I_{\text{D}}(\cdot)$. The proximal operator for the $\ell_1$-norm can be expressed as the solution to the following convex optimization program:
\begin{align}
\underset{z}{\text{minimize }} \lambda \|z\|_{\ell_1} + \left(\rho/2\right)\|z - x\|_{\ell_2}^2 
\end{align}
This problem has a closed-form solution given by the soft-thresholding operator $S_{1/\rho}(\cdot)$ \cite{Boyd2011a}, which takes the following form:
\begin{align}
x^* = S_{\lambda}(x) = \begin{cases}
x - \lambda \operatorname{sign}(x), & |x| > \lambda \\
0 & |x| \le \lambda
\end{cases}
\end{align}
Note that the $w_1^{(k+1)}$ FISTA update step is applied with $\lambda = t^{(k)}$.  The proximal operator for the quadratic error set can be written as the solution to the following optimization problem:
\begin{alignat}{3}
& \underset{z}{\text{minimize }}~~ & & \|z - x\|_{\ell_2}^2 \label{eq:ell2prox}\\
&\text{subject to } &  & \|z\|_{\ell_2}  \le \delta \nonumber
\end{alignat}
It is easy to show that Eq. \ref{eq:ell2prox} has the following close-form solution:
\begin{align}
x^* = \begin{cases}
x & \|x\|_{\ell_2} \le \delta \\
\left(\frac{\delta}{\|x\|_{\ell_2}}\right)x & \|x\|_{\ell_2} > \delta 
\end{cases}
\label{eq:simple_ell2_prox_sol}
\end{align}
The proximal operator for $I_{\text{box}}(\cdot)$ can be expressed as the solution to the following problem:
\begin{alignat}{3}
& \underset{z}{\text{minimize }}~~ & & \|z - x\|_{\ell_2}^2 \label{eq:boxprox}\\
&\text{subject to } &  & z \succeq 0_{N_t} \nonumber
\end{alignat}
which has the trivial closed-form solution:
\begin{align}
x^* = \operatorname{max}\left(x, 0\right)
\end{align}
where $\operatorname{max}$ operates on the individual elements of $x$. Finally, the proximal operator for $I_{\text{D}}(\cdot)$ can be expressed as the solution to the following problem:
\begin{alignat}{3}
& \underset{z}{\text{minimize }}~~ & & \|z - x\|_{\ell_2}^2 \label{eq:eqprox}\\
&\text{subject to } &  & Dz = 1_{N} \nonumber
\end{alignat}
This problem also has a closed-form solution, which after exploiting the structure $D$ can be expressed as:
\begin{align}
x^* = x + \frac{1}{R}\left(1_{N_t} - D^TDx\right)
\end{align}

\vspace{7pt}
\section{Numerical Results}
\label{sec:results}
\begin{figure}[h!]
\centering
\includegraphics[width=8cm, clip=true]{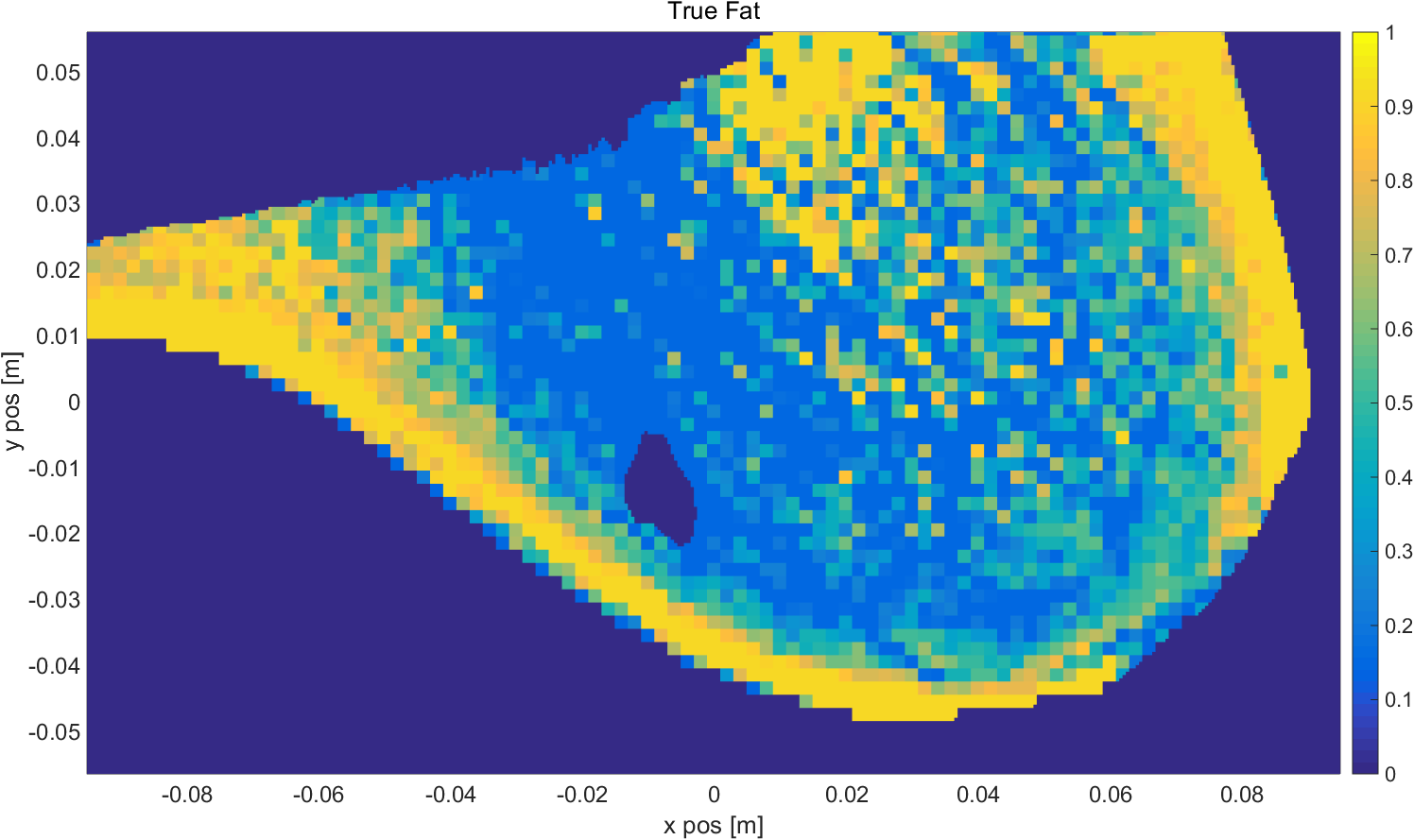}
    \caption{True LWC tissue mixture proportion.}
   \label{fig:fat_true}
\end{figure}

\begin{figure}[h!]
\centering
\includegraphics[width=8cm, clip=true]{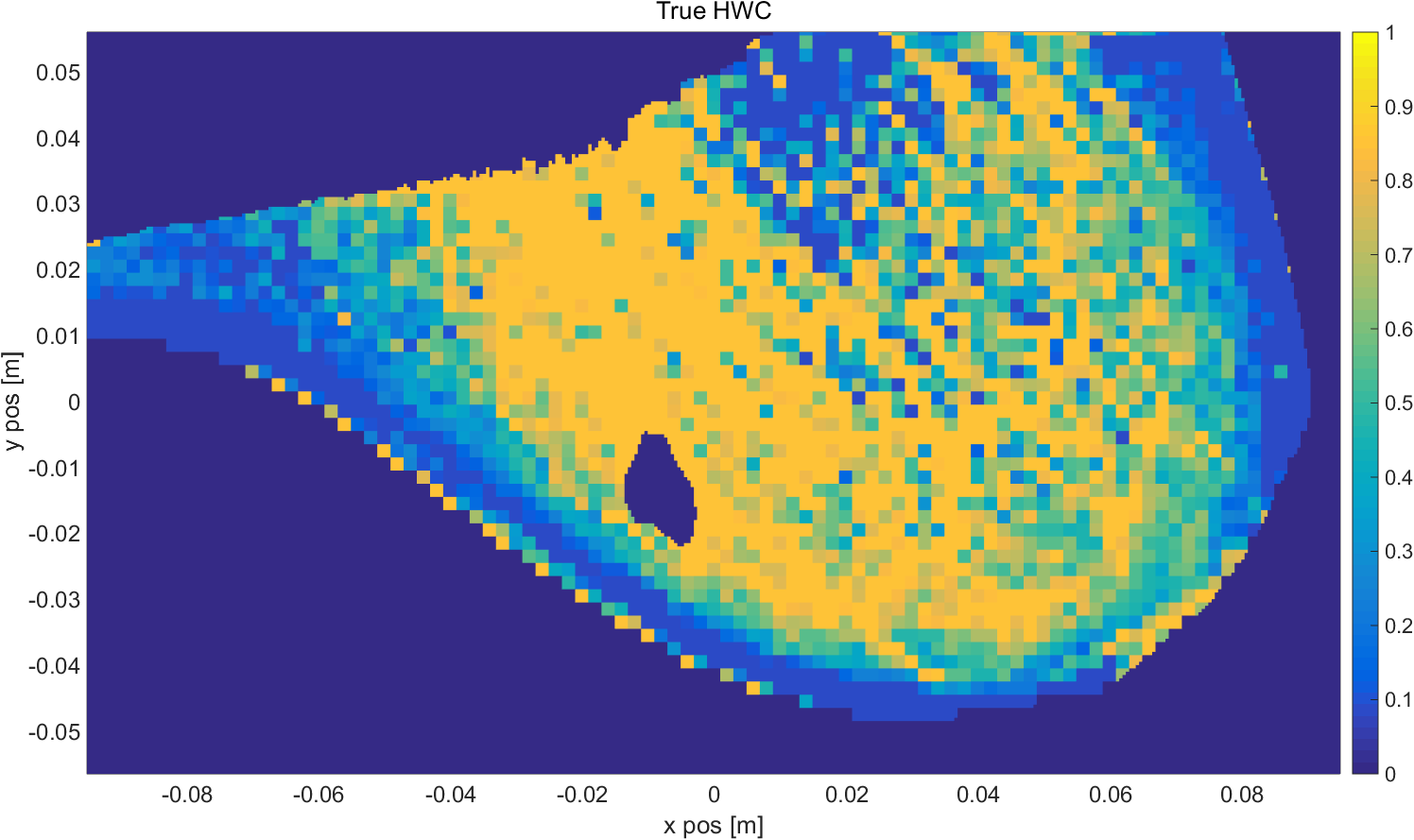}
    \caption{True HWC tissue mixture proportion.}
   \label{fig:hwc_true}
\end{figure}

\begin{figure}[h!]
\centering
\includegraphics[width=8cm, clip=true]{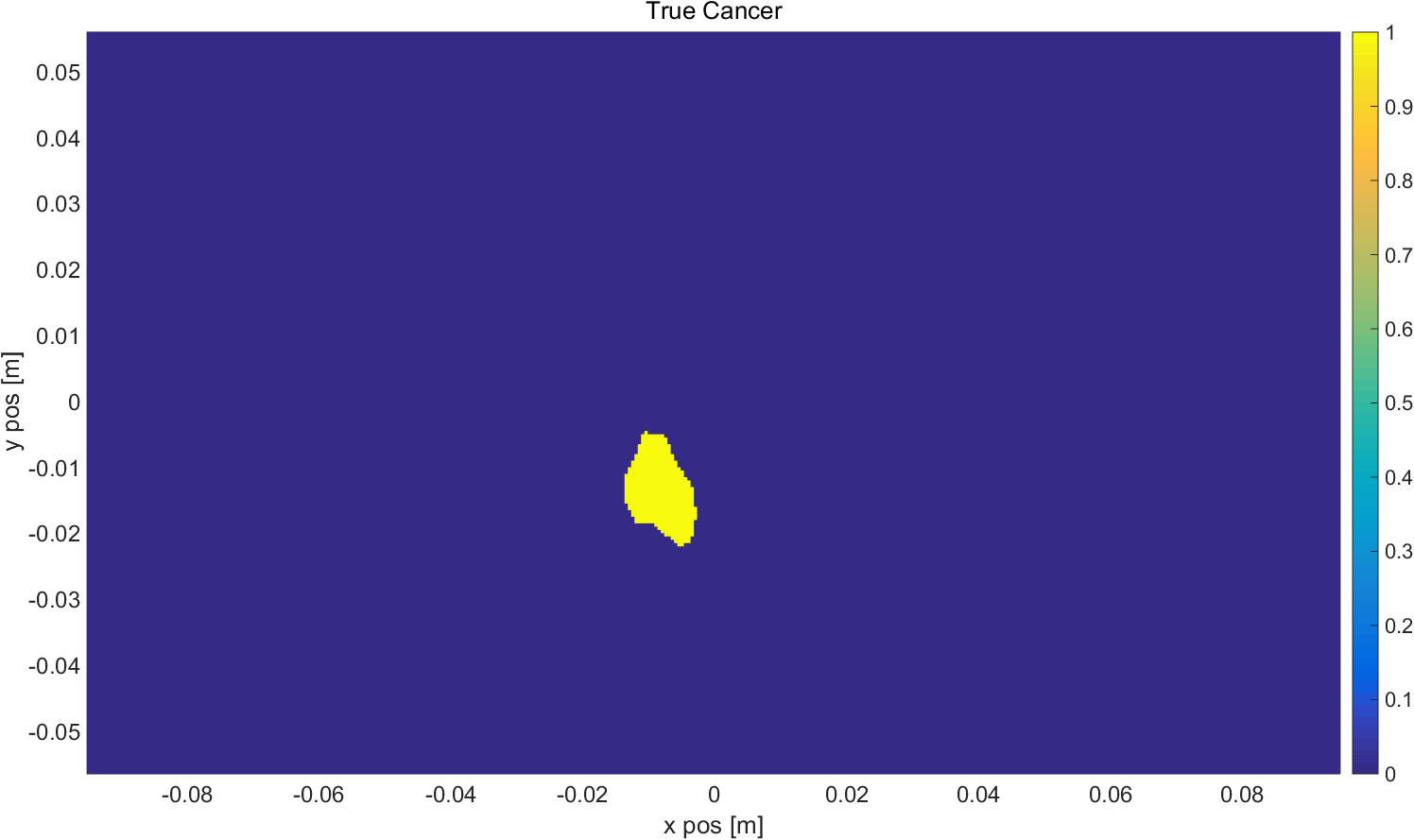}
    \caption{True cancerous tissue mixture proportion.}
   \label{fig:cnc_true}
\end{figure}

In this section, we assess the performance of the compressive sensing unmixing algorithm using simulated data. A 2D FDFD model \cite{Rappaport2001} was used in order to generate synthetic electric field measurements for two breast geometries, one with a cancerous lesion and one without. The two geometries had the same HWC and LWC tissue proportions at all locations except for that of the cancerous lesion. The baseline, healthy breast geometry was segmented from a 2D slice of an actual 3D DBT reconstruction. Figures \ref{fig:fat_true}, \ref{fig:hwc_true}, and \ref{fig:cnc_true} display the true mixture proportions for LWC tissue, HWC tissue, and cancerous tissue, respectively, of the unhealthy breast geometry.
In the numerical simulations, the breast geometries were excited by $17$ transmitting and receiving antennas operating in a multistatic configuration. Each transmitting antenna operated at $11$ frequencies linearly spaced from $500$ MHz to $1500$MHz, for a total of $3179$ complex measurements. Note that redundant measurements were used in the optimization routine. The healthy breast geometry simulations were used in order to generate the adjusted measurements $\hat{y}$ and to compute the Jacobian matrix $A$ required by the optimization procedure. The imaging region was constrained to $9654$ positions in the breast, where the grid size of each pixel was $2$mm. In order to consider the problem in the most ideal scenario possible, random noise was not added to the measurements. As a result, the measurements were only corrupted by noise introduced into the problem when it was linearized via the Born Approximation. This noise was estimated to have $12.5\%$ the energy of the adjusted measurement vector, i.e. $\eta \approx 0.125\|\hat{y}\|_{\ell_2}$. In addition, the difference between the measurements of the unhealthy breast, $y$, and the measurements of the healthy breast, $\tilde{y}=f(h(v_1,v_2,v_3))$, had approximately $12.69\%$ the energy of the adjusted measurement vector, i.e. $\|y - \tilde{y}\|_{\ell_2} \approx 0.1269\|\hat{y}\|_{\ell_2}$. As a result, the parameter $\delta$ in the optimization procedure can be no greater than $0.1269\|\hat{y}\|_{\ell_2}$, otherwise the optimal solution to the problem will be the initial proportions $v_1$, $v_2$, $v_3$. 

\begin{figure}[h!]
\centering
\includegraphics[width=8cm, clip=true]{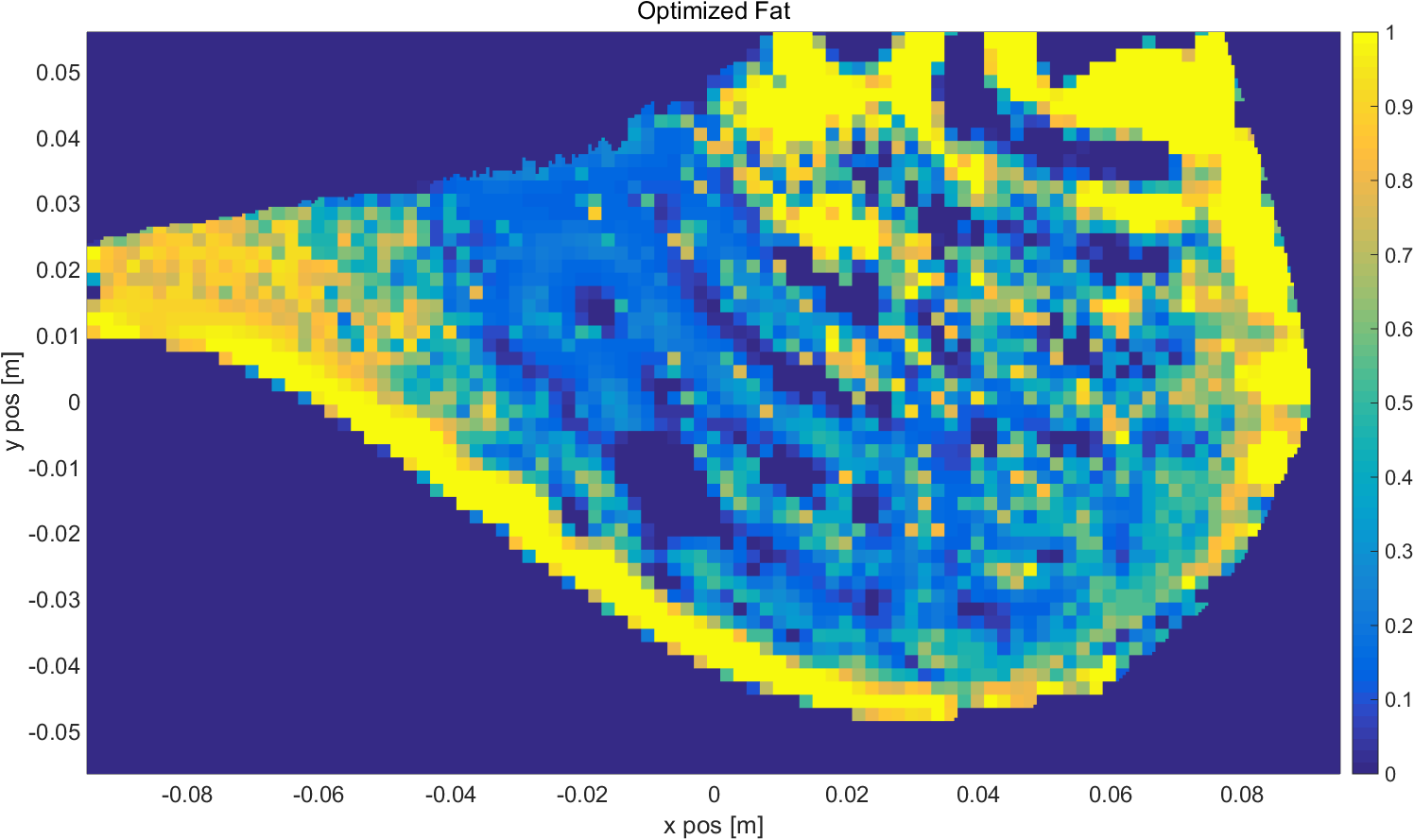}
    \caption{Estimated LWC tissue mixture proportion for $\delta = \|y\|_{\ell_2}/10000$.}
   \label{fig:fat_est_10000}
\end{figure}

\begin{figure}[h!]
\centering
\includegraphics[width=8cm, clip=true]{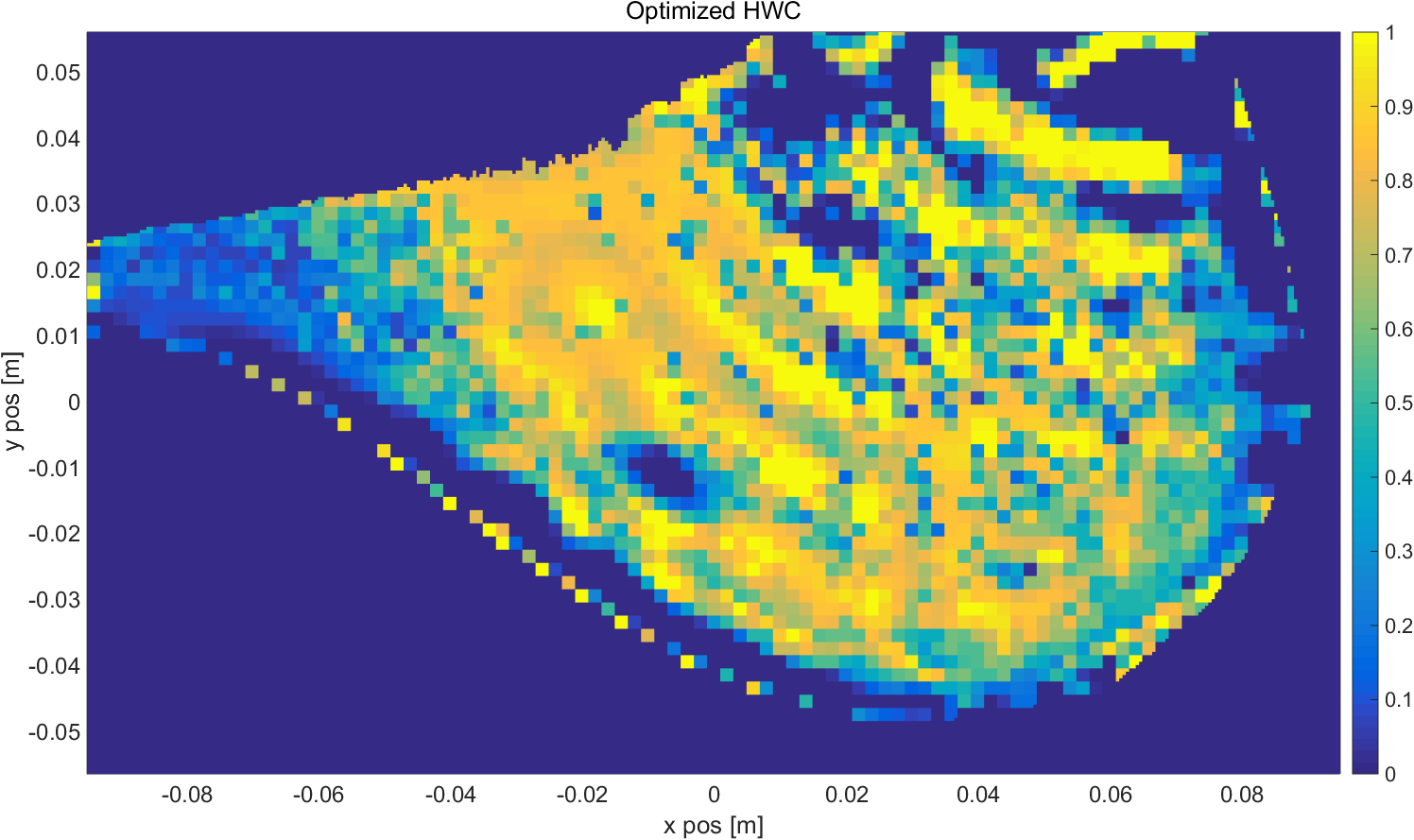}
    \caption{Estimated HWC tissue mixture proportion for $\delta = \|y\|_{\ell_2}/10000$.}
   \label{fig:hwc_est_10000}
\end{figure}

\begin{figure}[h!]
\centering
\includegraphics[width=8cm, clip=true]{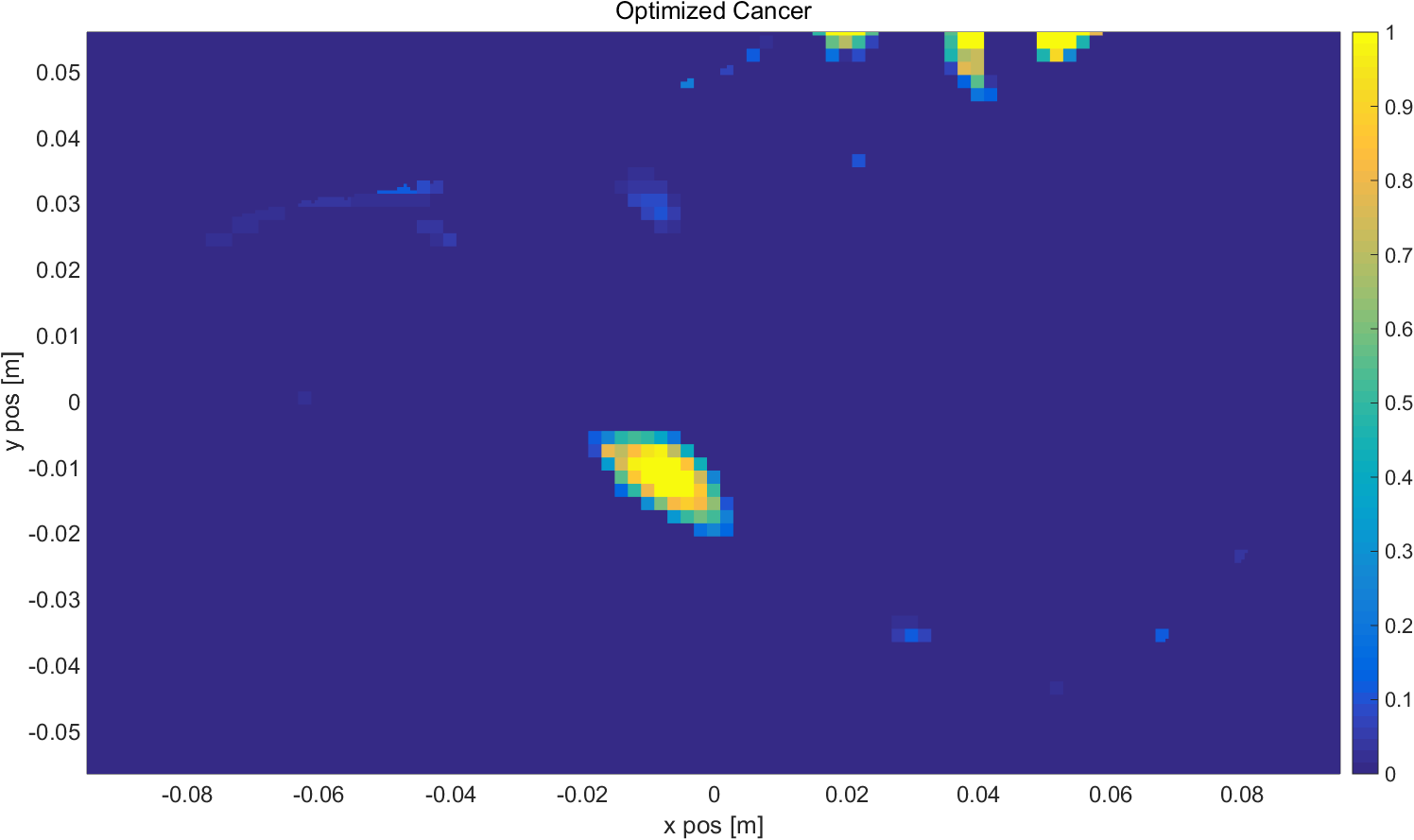}
    \caption{Estimated cancerous tissue mixture proportion for $\delta = \|y\|_{\ell_2}/10000$.}
   \label{fig:cnc_est_10000}
\end{figure}

\begin{figure}[h!]
\centering
\includegraphics[width=8cm, clip=true]{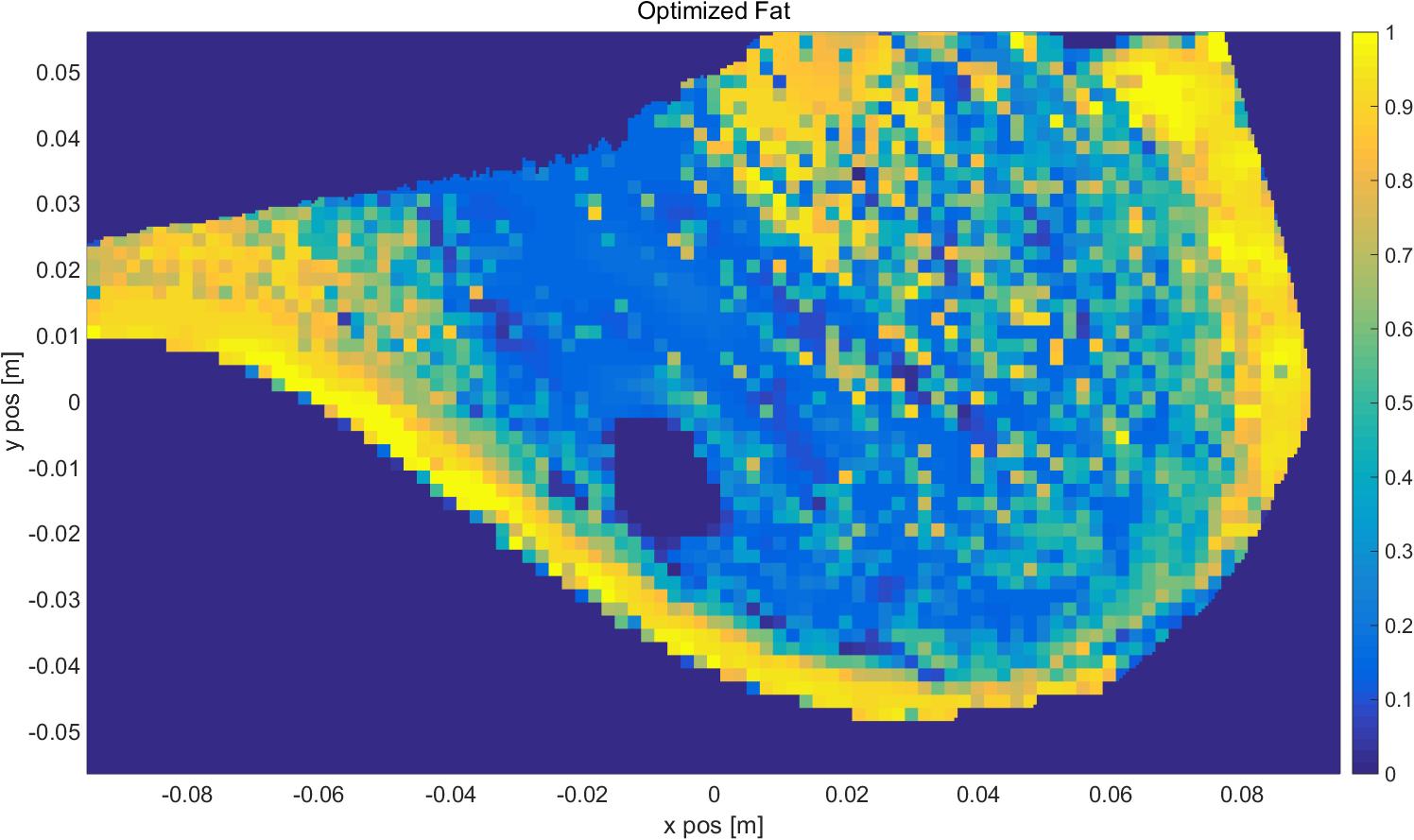}
    \caption{Estimated LWC tissue mixture proportion for $\delta = \|y\|_{\ell_2}/100$.}
   \label{fig:fat_est_100}
\end{figure}

\begin{figure}[h!]
\centering
\includegraphics[width=8cm, clip=true]{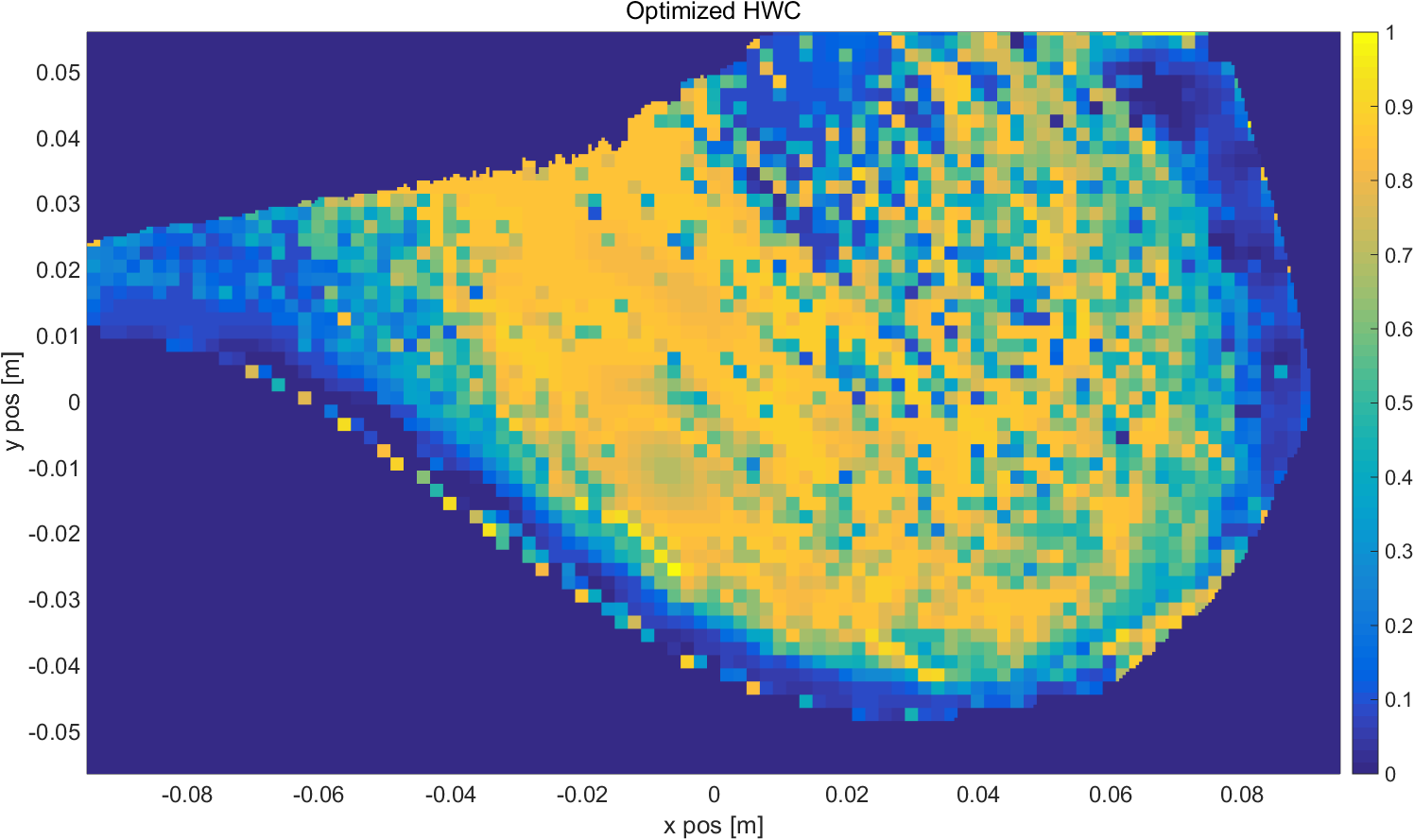}
    \caption{Estimated HWC tissue mixture proportion for $\delta = \|y\|_{\ell_2}/100$.}
   \label{fig:hwc_est_100}
\end{figure}

\begin{figure}[h!]
\centering
\includegraphics[width=8cm, clip=true]{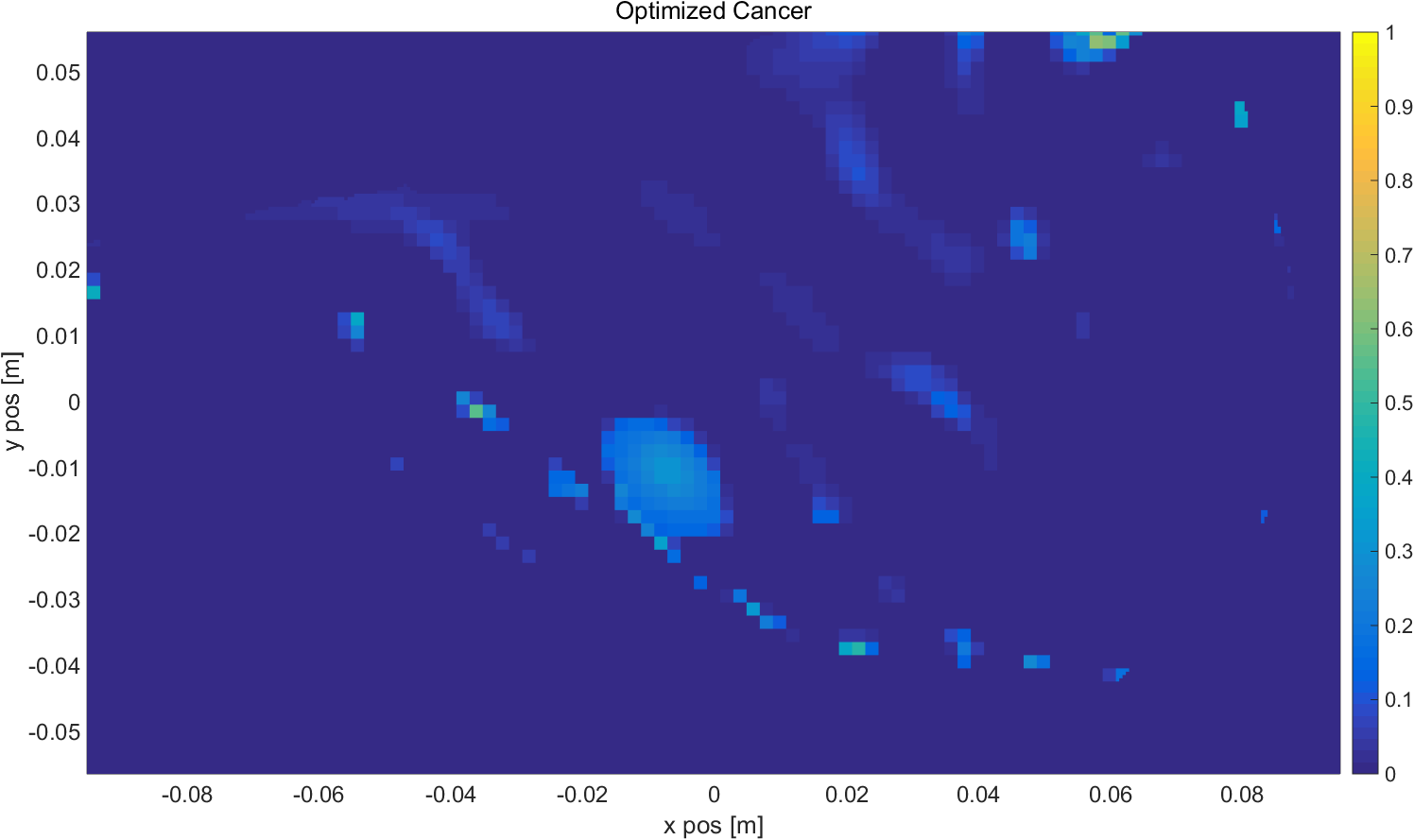}
    \caption{Estimated cancerous tissue mixture proportion for $\delta = \|y\|_{\ell_2}/100$.}
   \label{fig:cnc_est_100}
\end{figure}

\begin{figure}[h!]
\centering
\includegraphics[width=8cm, clip=true]{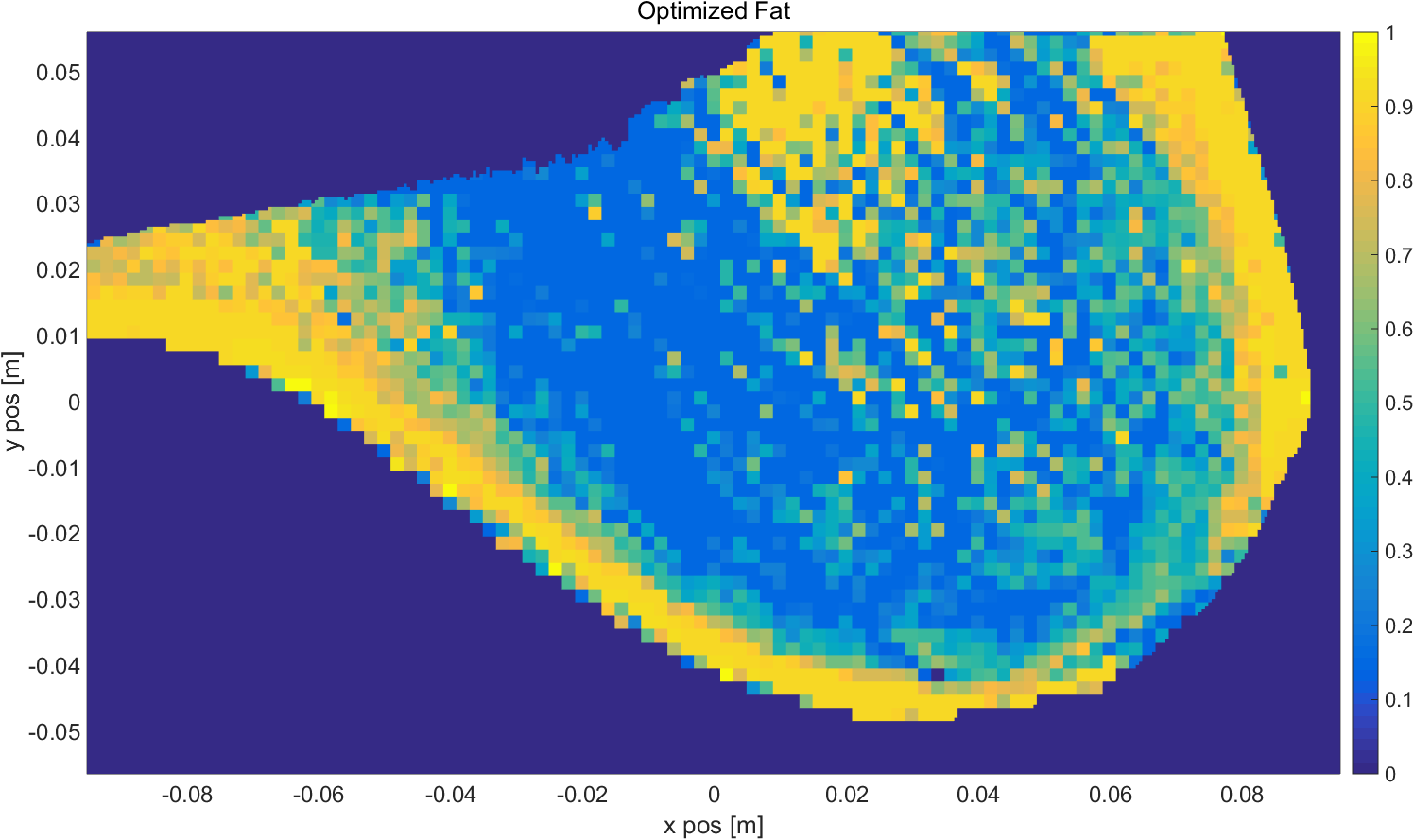}
    \caption{Estimated LWC tissue mixture proportion for $\delta = \|y\|_{\ell_2}/20$.}
   \label{fig:fat_est_20}
\end{figure}

\begin{figure}[h!]
\centering
\includegraphics[width=8cm, clip=true]{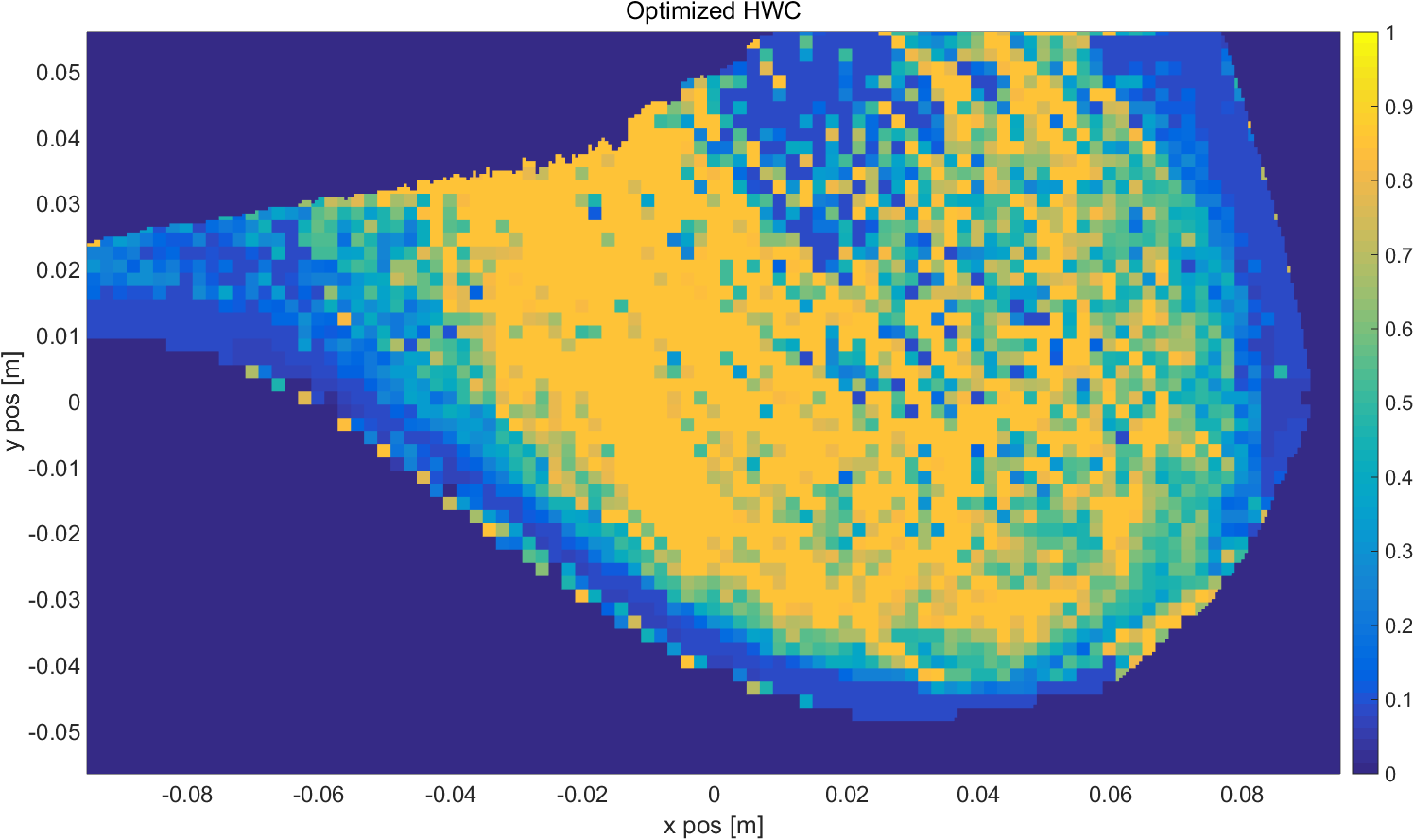}
    \caption{Estimated HWC tissue mixture proportion for $\delta = \|y\|_{\ell_2}/20$.}
   \label{fig:hwc_est_20}
\end{figure}

\begin{figure}[h!]
\centering
\includegraphics[width=8cm, clip=true]{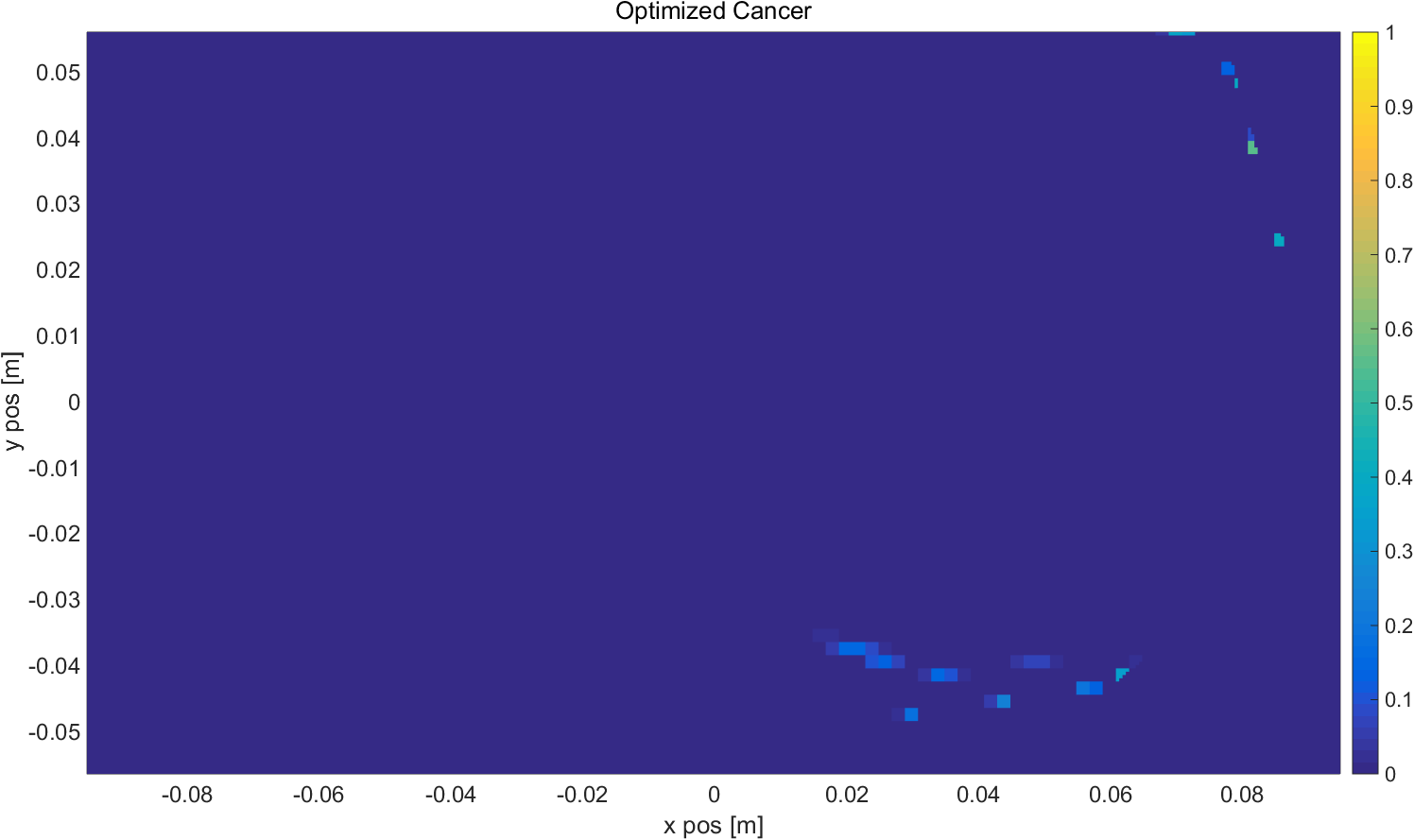}
    \caption{Estimated cancerous tissue mixture proportion for $\delta = \|y\|_{\ell_2}/20$.}
   \label{fig:cnc_est_20}
\end{figure}

Figures \ref{fig:fat_est_10000}, \ref{fig:hwc_est_10000}, and \ref{fig:cnc_est_10000} display the estimated mixture proportions for LWC, HWC, and cancerous tissue when $\delta = \|y\|_{\ell_2}/10000$ is used in Eq. \ref{eq:csunmix}. The mixture proportions are not exactly recovered, which is to be expected given that the true solution vector has an error of $0.125\|\hat{y}\|_{\ell_2}$ due to the Born Approximation. Nevertheless, the location of the cancerous lesion within Figure \ref{fig:cnc_est_10000} agrees with the ground truth image of Figure \ref{fig:cnc_true}. Figures \ref{fig:fat_est_100} - \ref{fig:cnc_est_20} display similar figures generated using $\delta = \|y\|_{\ell_2}/100$ and $\delta = \|y\|_{\ell_2}/20$. In the former, the cancerous lesion is located, albeit with a significantly reduced mixture proportion. In the latter, the cancerous lesion is not located at all. This analysis suggests that, in practice, the NRI sensor will need to have a large signal to noise ratio, and the forward model used to generate the adjusted measurements $\hat{y}$ and Jacobian matrix $A$ must be very accurate, in order for the unmixing procedure to be effective in a clinical setting.



\vspace{7pt}
\section{Conclusion}
\label{sec:conclusion}
This paper presents a novel signal processing methodology for detecting breast cancer using a hybrid DBT / NRI sensing system. Using the DBT reconstruction to generate prior distributions for the LWC and HWC tissue proportions within the breast, the new methodology is able to combine unmixing algorithms with compressive sensing theory in order to reconstruct the proportion of cancerous tissues within the breast. Numerical results indicate that both the measurement and modeling systems must be capable of generating high fidelity solutions in order for the unmixing approach to be feasible in a clinical setting. 

Although this paper only considered the NRI reconstruction problem, the use of unmixing methods in the inversion process opens up several exciting areas of further research. In particular, parameterizing the breast tissues in terms of LWC, HWC, and cancerous tissue proportions opens up the possibility for performing a joint inversion using several sensing modalities, i.e. DBT, NRI, and ultrasound. Indeed, if the baseline tissue proportions for the NRI reconstruction process are segmented directly from the DBT image, then it should be possible to formulate the DBT reconstruction process in terms of the mixture proportions as well. If done properly, a joint inversion method should be able to more accurately detect cancerous lesions than each of the single modality inversion methods applied by themselves. This is a topic of future research.


%
%



%
\vspace{7pt}

\bibliographystyle{IEEEtran}
\bibliography{./SICA-TA}

\end{document}